\documentclass[letterpaper, 10pt, conference]{ieeeconf}
\IEEEoverridecommandlockouts
\usepackage{cite}
\usepackage{amsmath,amssymb,amsfonts}
\usepackage{graphicx}
\usepackage{textcomp}
\usepackage{xcolor}
\usepackage[acronym, shortcuts, hyperfirst=false]{glossaries} 
\usepackage[draft]{hyperref}  
\usepackage[noabbrev]{cleveref}
\usepackage{algorithm}
\usepackage[noend]{algpseudocode}
\usepackage{caption}
\usepackage{subcaption}

\usepackage{todonotes}
\usepackage{booktabs}

\renewcommand{\baselinestretch}{0.98}

\definecolor{bluegreen}{HTML}{4488b2}
\definecolor{redpink}{HTML}{f4858d}
\definecolor{darkblue}{HTML}{4b50fc}

\definecolor{yellowgreen}{HTML}{98ff4d}
\definecolor{rust}{HTML}{a24443}
\definecolor{orange}{HTML}{f2a147}

\newacronym{drl}{DRL}{deep reinforcement learning}
\newacronym{pacmap}{PaCMAP}{Pairwise Controlled Manifold Approximation Projection}
\newacronym{rl}{RL}{reinforcement learning}
\newacronym{sac}{SAC}{Soft Actor-Critic}
\newacronym{ddpg}{DDPG}{Deep Deterministic Policy Gradient}
\newacronym{pca}{PCA}{Principal component analysis}
\newacronym{gan}{GAN}{Generative Adversarial Network}
\newacronym{dice}{DiCE}{Diverse Counterfactual Explanations}
\newacronym{lidar}{LiDAR}{Light Detection and Ranging}
\newacronym{xai}{XAI}{Explainable Artificial Intelligence}
\newacronym{cfe}{CFE}{Counterfactual Explanation}
\newacronym{ml}{ML}{Machine Learning}
\newacronym{ros}{ROS}{Robot Operating System}
\newacronym{ga}{GA}{Genetic algorithm}

\begin{document}

\title{Realistic Counterfactual Explanations for Machine Learning-Controlled Mobile Robots using 2D LiDAR}

\author{Sindre Benjamin Remman and Anastasios M. Lekkas
\thanks{The authors are with the Department of Engineering Cybernetics,
        Norwegian University of Science and Technology (NTNU), Trondheim, Norway
        {\tt\small \{sindre.b.remman, anastasios.lekkas\}@ntnu.no}}%
}

\maketitle

\begin{abstract}
This paper presents a novel method for generating realistic counterfactual explanations (CFEs) in machine learning (ML)-based control for mobile robots using 2D LiDAR. ML models, especially artificial neural networks (ANNs), can provide advanced decision-making and control capabilities by learning from data. However, they often function as black boxes, making it challenging to interpret them. This is especially a problem in safety-critical control applications. To generate realistic CFEs, we parameterize the LiDAR space with simple shapes such as circles and rectangles, whose parameters are chosen by a genetic algorithm, and the configurations are transformed into LiDAR data by raycasting. Our model-agnostic approach generates CFEs in the form of synthetic LiDAR data that resembles a base LiDAR state but is modified to produce a pre-defined ML model control output based on a query from the user. We demonstrate our method on a mobile robot, the TurtleBot3, controlled using deep reinforcement learning (DRL) in real-world and simulated scenarios. Our method generates logical and realistic CFEs, which helps to interpret the DRL agent's decision making. This paper contributes towards advancing explainable AI in mobile robotics, and our method could be a tool for understanding, debugging, and improving ML-based autonomous control. 
\end{abstract}


\section{Introduction}

With the rapid progress of \ac{ml} applications in robotics \cite{soori2023artificial}, it is crucial to ensure that the decision-making by \ac{ml} models is transparent \cite{wachter2017transparent}. In safety-critical control tasks, model interpretability can help us understand and trust the control decisions of robots controlled using neural networks, which often function like black boxes. \glspl{cfe}, also called counterfactuals, produce contrastive explanations by showing how small input changes to an \ac{ml} model can lead to different outputs. \glspl{cfe} are a promising way to increase interpretability through "what if" scenarios \cite{wachter2017counterfactual}.

\ac{cfe}s have been used in robotics, for instance, in \cite{smith2020counterfactual}, which uses them to assess robot control robustness. The authors train a generative model to create small and realistic modifications to a base image to produce user-defined effects in an image-based controller, thus producing \ac{cfe}s. These \ac{cfe}s are then employed as a measurement of the robustness of the controller, i.e., if the \ac{cfe}s have to be very different from the base image to produce significantly different effects in the controller, the controller is likely robust. Another application is presented in \cite{gjaerum2023real}, where real-time \ac{cfe}s are generated for robotic systems with multiple continuous outputs. The authors use linear model trees to produce the \ac{cfe}s and demonstrate that while the method might find infeasible \ac{cfe}s if the input features are not independent, they can perform feature engineering to ensure the input features are independent. Although not robotics, \cite{samadi2024safe} propose a novel solution for generating \ac{cfe}s for image-input-based \ac{drl} agents. They use a saliency map to discover the most important input pixels and use this as input to a generative model that generates realistic \ac{cfe}s. 

While there has been work on \ac{cfe}s in robotics and \ac{drl}, to our knowledge, no existing methods deal with \ac{cfe}s for high-dimensional LiDAR data. In this paper, we demonstrate how applying \ac{dice}, an existing \ac{cfe} generation method to LiDAR data can lead to noisy and unrealistic results due to the correlations often present between neighboring readings in LiDAR data. We propose a model-agnostic approach for generating realistic \glspl{cfe} tailored for mobile robots with 2D LiDAR. Although demonstrated here with an ML-based controller, this method could also be used for traditional control algorithms because of its model-agnostic workings. By parameterizing the LiDAR space with geometric shapes, we ensure the \ac{cfe}s resemble realistic environments. Our method can, therefore, give a true and intuitive understanding of a model's decision-making. We test our method on a TurtleBot3 with a \ac{drl} policy in both simulated and real-world settings. Our method surpasses existing techniques like \ac{dice} in interpretability and similarity to real data when applied to 2D LiDAR data.

Our contributions include: 
\begin{itemize} 
\item A novel, model-agnostic method for generating spatially realistic \ac{cfe}s for robots with 2D LiDAR, which applies a genetic algorithm approach inspired by \ac{dice}, with significant modifications to handle the spatial structure of LiDAR data.
\item A comparison against \ac{dice}, demonstrating enhanced interpretability and realism for 2D LiDAR data. 
\item Real-world tests on a TurtleBot3 controlled by \ac{drl}, demonstrating the method’s applicability to control tasks and its effectiveness with real-world data.
\item An experiment using generated LiDAR-like data to investigate the \ac{ml} model’s behavior preferences, demonstrating how our methodology can improve understanding of \ac{ml} models.
\end{itemize}

\section{Background and Theory}\label{sec:background}

This section introduces \ac{cfe}s and genetic algorithms, which are important concepts in our approach. \ac{cfe}s help interpret ML models by showing how small input changes affect outcomes based on user queries. Genetic algorithms are the optimization methods that we use to generate our \ac{cfe}s. We discuss how the methodology uses these concepts in Section~\ref{sec:methodology}.

\subsection{Counterfactual Explanations}

Effective explanations are often \textit{contrastive}, and answer questions like, "Why not this instead?" rather than simply "Why?" \cite{miller2019explanation, molnar2022}. In \ac{ml}, \ac{cfe}s are particularly suited for this since they show how changes in input can produce different, user-defined outcomes. For instance, in a model classifying images of cats and dogs, a \ac{cfe} could explain how an image of a cat would need to change to be labeled as a dog.

In robotics, \ac{cfe}s can answer questions like "What changes could make the model choose action \textit{B} over action \textit{A}?" The continuous action spaces in robotics make it difficult to find precise numerical outputs, so it is more relevant to ask, "What changes could make the model choose an action within bounds \textit{B} rather than a specific action \textit{A}?"

Inspired by DiCE's approach \cite{dice2020github}, we define CFEs for continuous outputs as follows: Let $\mathbf{x} \in \mathbb{R}^n$ represent the input space and $f: \mathbb{R}^n \rightarrow \mathbb{R}^m$ be the \ac{ml} model, mapping inputs to actions $\mathbf{a} = f(\mathbf{x}) \in \mathbb{R}^m$. For a target action bound $\mathbf{B} \subset \mathbb{R}^m$, a \ac{cfe} $\mathbf{x}_{\text{cf}}$ is an adjusted input such that:

$$
f(\mathbf{x}_{\text{cf}}) \in \mathbf{B} \quad \text{and} \quad \lVert \mathbf{x} - \mathbf{x}_{\text{cf}} \rVert \leq \epsilon,
$$

where $f(\mathbf{x}_{\text{cf}})$ is inside the target bounds $\mathbf{B}$ with minimal deviation $\epsilon$ from the original input $\mathbf{x}$.

\subsection{Genetic Algorithms}

\glspl{ga}, introduced by Holland in \cite{holland1975adaptation}, are optimization methods inspired by natural selection, suited to non-linear, non-convex problems \cite{katoch2021review}. Our approach uses a \ac{ga} to optimize candidate solutions, or individuals, for generating realistic LiDAR \ac{cfe}s. The whole set of candidate solutions is called the population. It consists of individuals structured as chromosomes, composed of genes that encode specific parameters and are evaluated by a fitness function, also called an objective function.

To evolve the population, a \ac{ga} uses three main operators \cite{katoch2021review}:
\begin{itemize}
    \item \textit{Crossover}: Generates offspring by combining genes from two or more parents.
    \item \textit{Mutation}: Maintains diversity between generations by introducing random changes. 
    \item \textit{Elitism}: Keeps the best solutions between generations to have steady progress.
\end{itemize}

The \ac{ga} iterates through several steps: initialization of a random population, evaluation of each individual's fitness, selection for crossover and mutation, and replacement to form the next generation. Elitism ensures that high-quality solutions are preserved and that solutions are steadily progressing. This process continues until a stopping criterion is met, such as reaching a maximum number of generations or achieving a certain fitness score.

\section{Methodology}\label{sec:methodology}

\subsection{Generating Realistic LiDAR Counterfactuals}

In our algorithm, detailed below, we use \textit{real} and \textit{virtual} LiDAR data. Real LiDAR data consists of sensor readings, the distances from the LiDAR to the closest objects at certain angle intervals 360 degrees around the sensor, collected directly from the environment. On the other hand, virtual LiDAR data is artificially generated to simulate hypothetical obstacles and configurations. This virtual data forms the basis of our \ac{cfe}s, showing how modified sensor inputs could affect the model’s decisions.

Generating realistic \ac{cfe}s from LiDAR data is challenging due to spatial correlations. For instance, consecutive LiDAR readings may reflect the same object. \ac{cfe}s generated without considering these correlations tend to resemble noise, as we demonstrate with the \ac{dice} method in \Cref{subsec:dice_comparison}.

\begin{figure}
    \centering
    \includegraphics[width=1.0\linewidth]{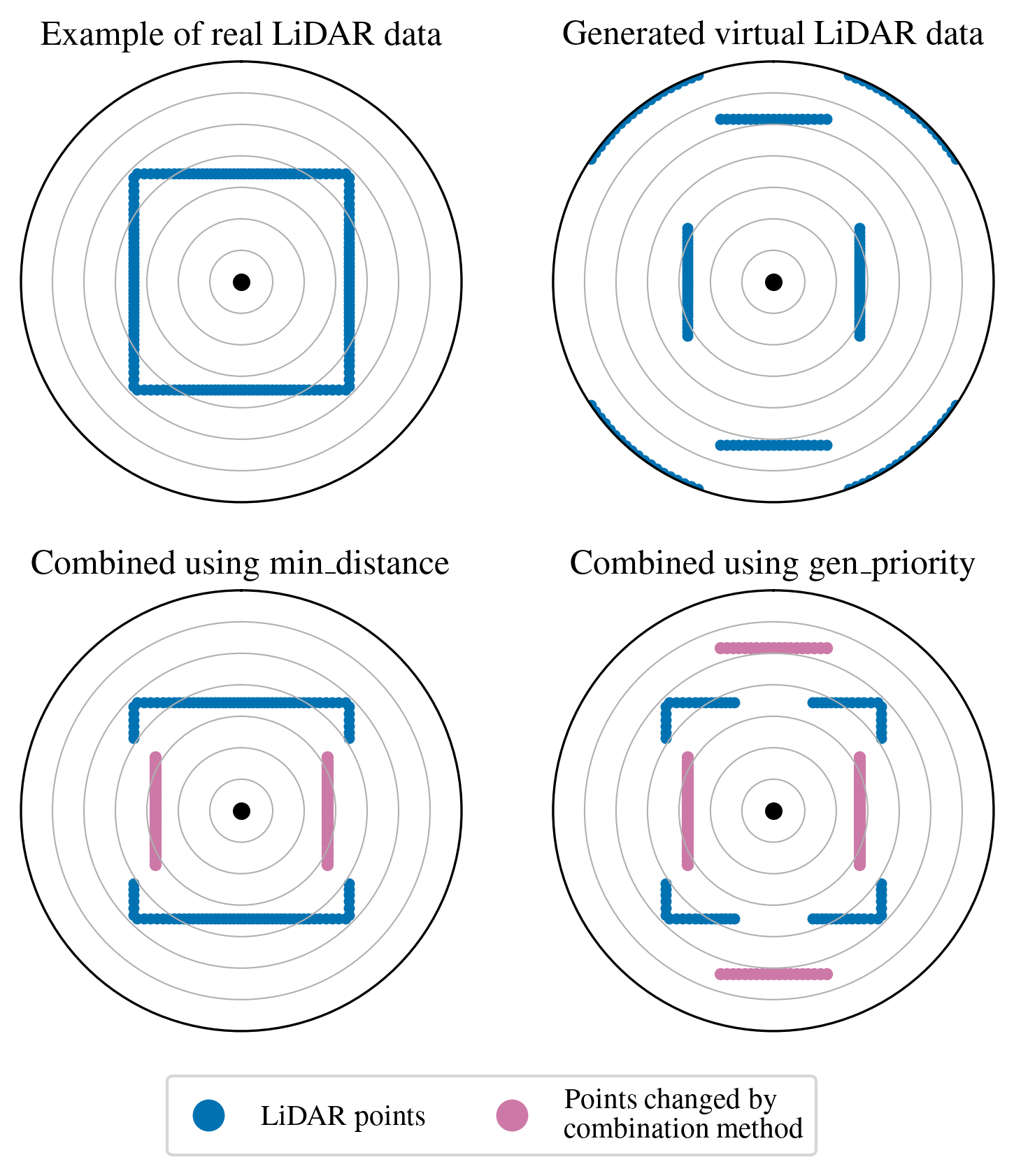}
    \caption{Example illustrating the two combination methods. The top left shows the base LiDAR state with four walls around the sensor, while the top right shows the generated virtual obstacles. The lower left shows the result of combining these data with \textit{min\_distance}, which keeps the nearest obstacles at each reading, and the lower right shows \textit{gen\_priority}, which prioritizes the virtual obstacles. The dot at the center marks the origin of the LiDAR sensor.}
\label{fig:combination_methods_example}
\end{figure}

Our methodology generates LiDAR \ac{cfe}s by modifying sensor readings to illustrate how small and realistic changes can lead to different model outputs. In our case study, shown in \Cref{subsec:casestudy}, the ML model, $f$, selects linear and angular velocities directly from LiDAR data. To generate realistic \ac{cfe}s, we parameterize the LiDAR space with basic shapes (e.g., circles, squares) and transform these into virtual LiDAR data by raycasting to the shapes' boundaries, using the Shapely package \cite{shapely}. Once generated, these virtual LiDAR points are combined with the original LiDAR state using two combination methods, illustrated in \Cref{fig:combination_methods_example} and explained below.

Let $ \mathbf{b} $ and $ \mathbf{c} $ represent vectors of LiDAR readings in meters, where:
\begin{itemize}
    \item $ \mathbf{b} = [b_1, b_2, \dots, b_n] $ denotes the baseline LiDAR readings, and
    \item $ \mathbf{c} = [c_1, c_2, \dots, c_n] $ denotes the generated LiDAR readings corresponding to the obstacles.
\end{itemize}

The first combination method, which we call \textit{min\_distance}, computes a new vector $ \mathbf{d} $ by taking the minimum value for each entry in $ \mathbf{b} $ and $ \mathbf{c} $, which element-wise is:
\begin{equation}\label{eq:min_distance}
d_i = \min(b_i, c_i) \quad \text{for } i = 1, 2, \dots, n.
\end{equation}

The second combination method, called \textit{gen\_priority}, prioritizes the generated readings, $\mathbf{c}$, whenever they fall within a specified threshold distance, $D_{\text{max}}$, which represents the maximum LiDAR range. This approach results in combined LiDAR readings that match the generated readings when a LiDAR ray would encounter a generated obstacle and defaults to baseline readings otherwise. This is defined element-wise as:

\begin{equation}\label{eq:prio}
d_i = 
\begin{cases}
c_i & \text{if } c_i < D_{\text{max}}, \\
b_i & \text{otherwise}.
\end{cases}\quad \text{for } i = 1, 2, \dots, n.
\end{equation}

Each of the two combination methods provides a different approach to combining baseline and generated LiDAR data to generate a LiDAR state that serves as the \ac{cfe} instance. 
The \textit{min\_distance} combination method represents what would happen if we placed the generated obstacles in the environment without changing anything else, which can make the \ac{cfe}s more actionable. On the other hand, \textit{gen\_priority} gives the algorithm more freedom by letting it make larger changes from the base state. 

The methodology's main challenge is placing the virtual obstacles so that a specific change in the \ac{ml} model’s output is achieved. Since raycasting and \ac{ml} models, such as neural networks, are generally non-linear and non-convex, we use a \ac{ga} rather than gradient-based optimization. To implement the \ac{ga}, we use the PyGAD Python package \cite{gad2023pygad} with parameters shown in \Cref{tab:pygad_ga_parameters}, where we use a mutation rate of 20\% to increase the chance of discovering diverse solutions. Our full implementation is available in \cite{remman2024realistic}.

\begin{table}
    \centering
    \caption{Genetic Algorithm Parameters for PyGAD}
    \begin{tabular}{ll}
        \toprule
        \textbf{Parameter} & \textbf{Value/Description} \\ 
        \midrule
        Number of Generations & 100 \\ 
        Number of Parents Mating & 10 \\ 
        Solutions per Population & 100 \\ 
        Parent Selection Type & Tournament \\ 
        Keep Parents & 10 \\ 
        Crossover Type & Single-point \\ 
        Mutation Type & Random \\ 
        Mutation Percent Genes & 20\% \\ 
        Stop Criteria & \{"saturate\_10", "reach\_0"\} \\ 
        \bottomrule
    \end{tabular}
    \label{tab:pygad_ga_parameters}
\end{table}

We define the gene space for the GA to encode each virtual obstacle using six genes, representing obstacle type, circle or rectangle here (one gene), position and orientation (three genes), and size (two genes). For circles, we ignore the orientation gene and one of the size genes. Each gene value is normalized between 0 and 1. If we have $o_n$ obstacles to generate, the gene space is then of size $o_n \times 6$.

Each chromosome, representing a specific configuration of obstacles, is evaluated by the fitness function shown in \Cref{alg:fitness_func} to calculate how well each candidate fulfills the requirements. This fitness function uses two loss functions, inspired by DiCE’s approach: hinge loss, $y_{\text{loss}}$, and proximity loss, $p_{\text{loss}}$:

\begin{equation}\label{eq:hinge_loss}
y_{\text{loss}} = 
\begin{cases} 
0, & \text{if } B_0 \leq \hat{y} \leq B_1, \\
\min \left( | \hat{y} - B_0 |, | \hat{y} - B_1 | \right), & \text{otherwise}.
\end{cases}
\end{equation}

\begin{itemize}
    \item $ y_{\text{loss}} $: Hinge loss
    \item $ \hat{y} $: Output of the \ac{ml} model
    \item $ B_0 $ and $ B_1 $: Lower and upper bounds of the desired range for $ \hat{y} $.
\end{itemize}

\begin{equation}\label{eq:prox_loss}
    p_{\text{loss}} = -\sum_{i=1}^{n} \left| d_i - b_i \right|
\end{equation}

\begin{itemize}
    \item $ d_i $ represents each element in the model input (i.e., the generated LiDAR data),
    \item $ b_i $ is the corresponding element in the baseline state,
    \item $ n $ is the dimension of the state.
\end{itemize}

The hinge loss encourages the output of the \ac{ml} model to be within the user-defined bounds, and the proximity loss encourages the generated solution's LiDAR representation to be close to the base state.

After running the \ac{ga} and receiving the solutions, we only need to decode and combine the solutions with the base LiDAR state using the selected combination function to receive our \ac{cfe}s. 

\begin{algorithm}
\caption{Realistic LiDAR \ac{cfe} Fitness Function}\label{alg:fitness_func}
\begin{algorithmic}[1]
\Require{Solution $s$, Baseline data $\mathbf{b}$, Output bounds $B$, Combination function $F$, \ac{ml} model $P$, $y_{\text{loss}}$ weight $\lambda_y$, $p_{\text{loss}}$ weight $\lambda_p$, Min distance from LiDAR $d_{min}$}
\Ensure{Fitness score $f(s)$}
\State Decode solution $s$ into spatial parameters for each obstacles, including type, size, position, and orientation.
\State \textbf{If} any obstacle overlap a circle with radius $d_{min}$ centered at the LiDAR origin, \textbf{then return} $f(s) \gets -\infty$
\State Generate LiDAR data $\mathbf{c}$ from the generated obstacles using raycasting.
\State Combine baseline $\mathbf{b}$ and generated $\mathbf{c}$ to obtain combined LiDAR state $\mathbf{d}$ using $F$.

\State \textbf{Initialize} $f(s) \gets 0$  \Comment{Initialize fitness score to zero}

\State Input combined state $\mathbf{d}$ to model $P$ to obtain output $\mathbf{y}$.
\State Calculate $y_{i, \text{loss}}$ for each $y_i$ in $\mathbf{y}$ using \Cref{eq:hinge_loss}.
\State $f(s) \gets f(s) - \sum_{i} \lambda_y \cdot y_{i, \text{loss}}$.
\State Calculate $p_{\text{loss}}$ between $\mathbf{d}$ and $\mathbf{b}$ using \Cref{eq:prox_loss}.
\State Normalize $p_{\text{loss}}$ over the LiDAR dimension.
\State $f(s) \gets f(s) - \lambda_p \cdot p_{\text{loss}}$.

\State \Return $f(s)$
\end{algorithmic}
\end{algorithm}

\subsection{TurtleBot3 case study}\label{subsec:casestudy}

\begin{figure}
    \centering
    \begin{subfigure}{0.45\linewidth}
        \centering
        \includegraphics[width=\linewidth]{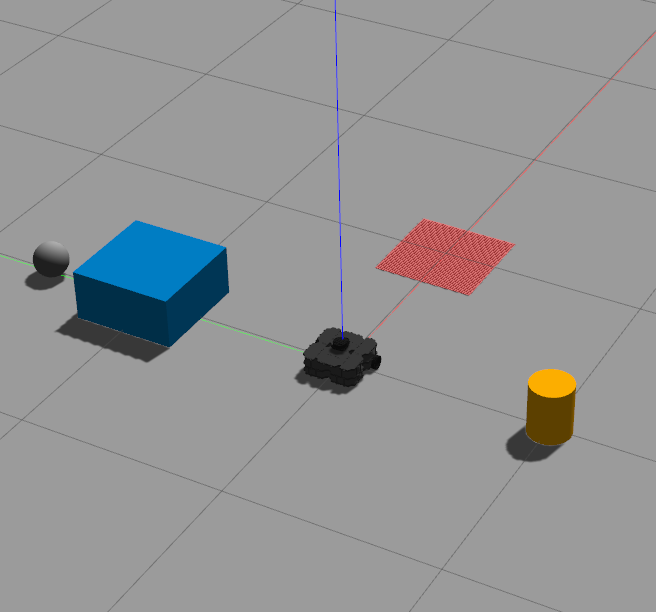} 
        \caption{Example Gazebo Scene}
    \end{subfigure}
    \hfill
    \begin{subfigure}{0.45\linewidth}
        \centering
        \includegraphics[width=\linewidth]{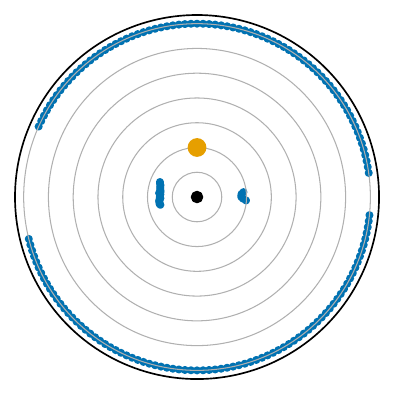} 
        \caption{Example Lidar Data Plot}
        \label{subfig:gazebo_lidar_comparison_lidar}
    \end{subfigure}

    \caption{Example of a Gazebo scene and the corresponding LiDAR plot. The goal in front of the TurtleBot3 is shown as a red square in Gazebo and an orange circle in the LiDAR plot. The TurtleBot3 is represented by the black dot in the middle of the plot, the blue square to the left of the TurtleBot3 is shown to the left of the TurtleBot3 in the plot, and vice versa with the orange cylinder to the right. The grey sphere is not visible in \Cref{subfig:gazebo_lidar_comparison_lidar} since it is behind the blue cube and not visible from the TurtleBot3's point of view.}
    \label{fig:gazebo_lidar_comparison}
\end{figure}

\begin{figure}
    \centering
    \begin{subfigure}{0.45\linewidth}
        \centering
        \includegraphics[ trim=0cm 18cm 0cm 10cm, clip,width=\linewidth]{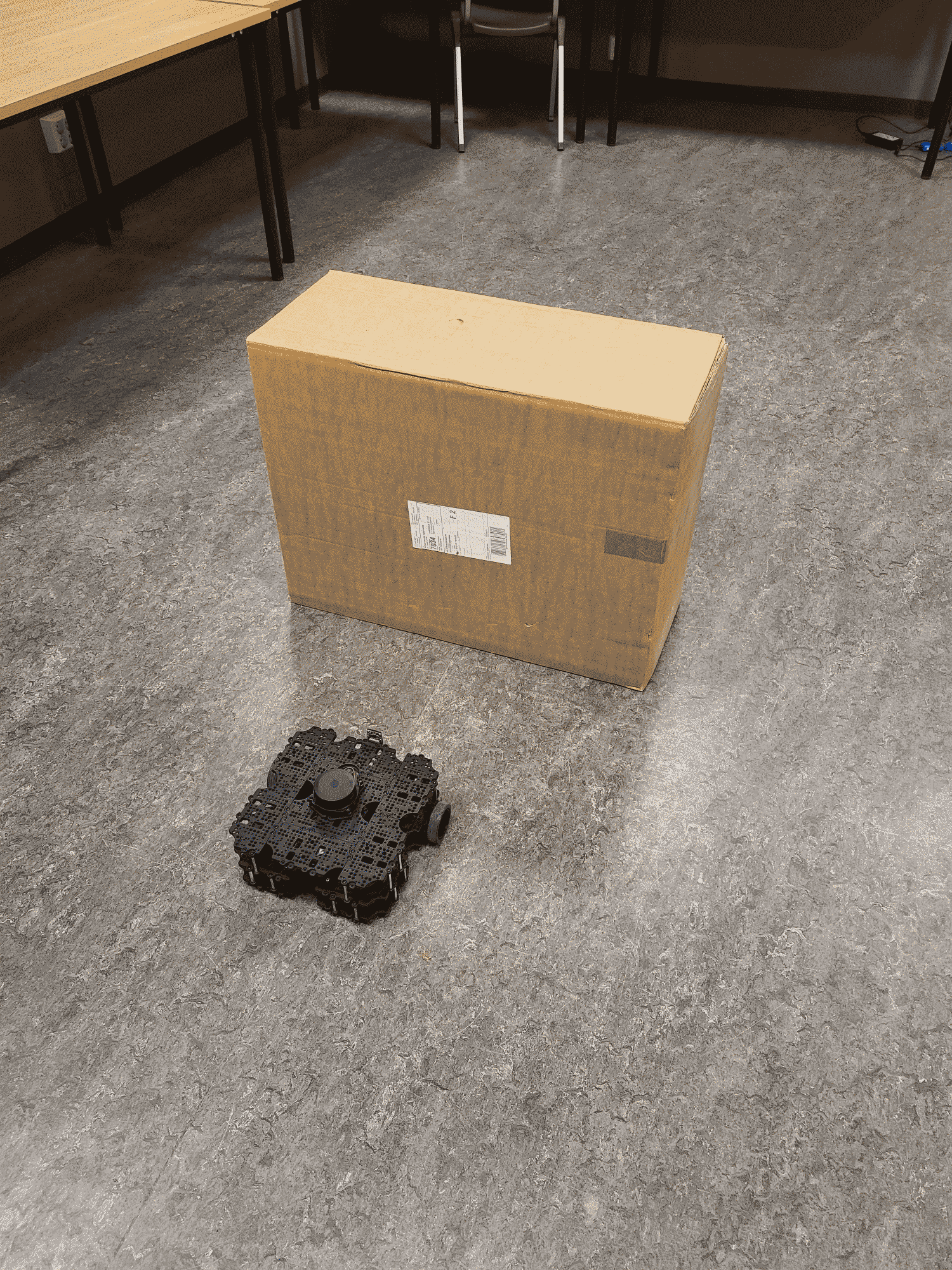}
        \caption{Initial state of Case 2.}
    \end{subfigure}
    \hfill
    \begin{subfigure}{0.45\linewidth}
        \centering
        \includegraphics[ trim=0cm 18cm 0cm 10cm, clip,width=\linewidth]{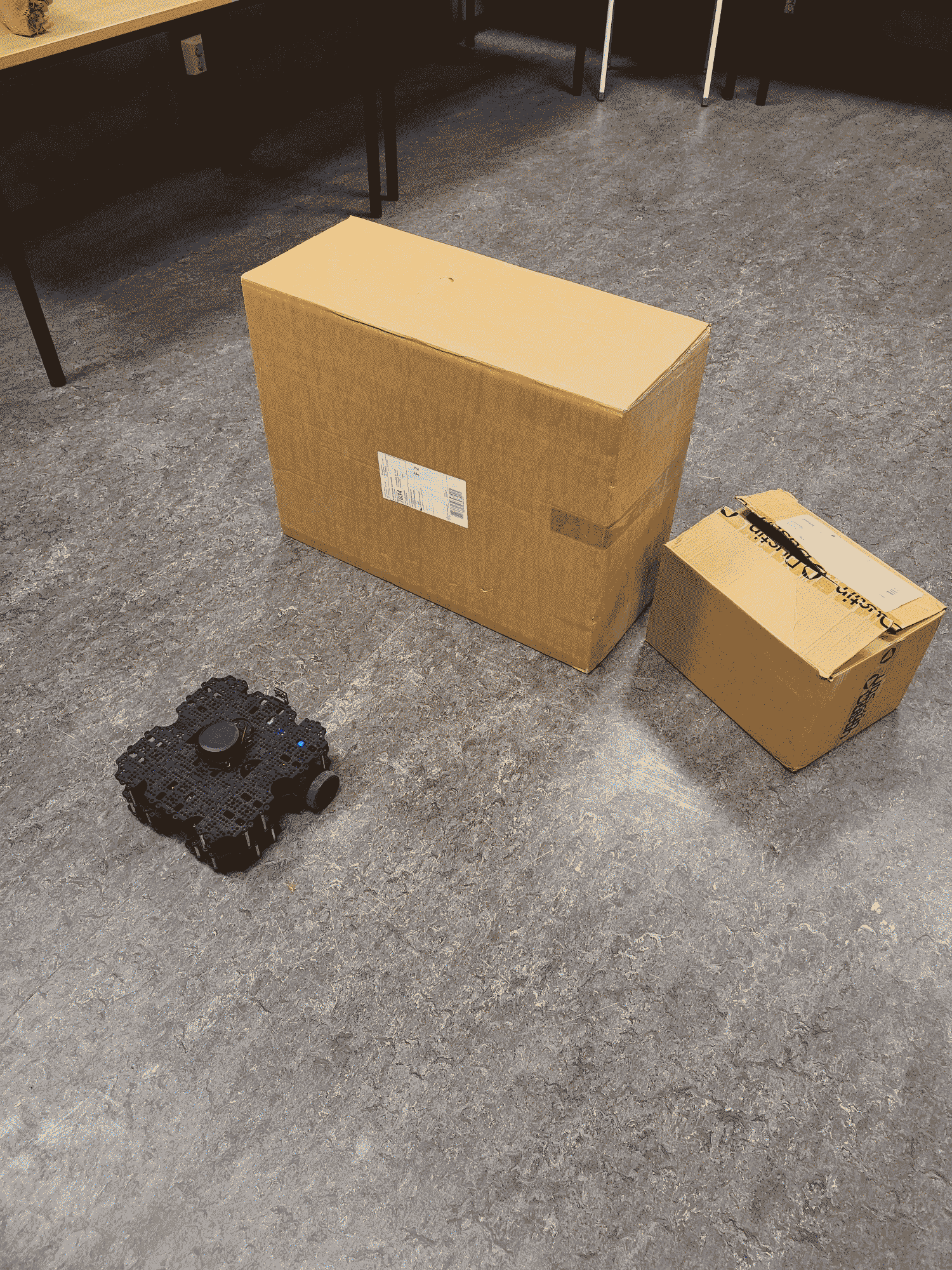}
        \caption{Case 2 after applying \ac{cfe}.}
    \end{subfigure}

    \caption{Photos of the real-world setup for Case 2, before and after actualizing the \ac{cfe}, discussed in \Cref{subsubsec:case_2}.}
    \label{fig:case_2_real_world}
\end{figure}

To evaluate our algorithm, we use the TurtleBot3\footnote{https://emanual.robotis.com/docs/en/platform/turtlebot3/overview/}, a low-cost, versatile mobile robot platform widely used in research and education. It has a 360-degree 2D LiDAR and is also integrated with \ac{ros}, which streamlines the controlling of the mobile robot. \Cref{fig:gazebo_lidar_comparison} shows the TurtleBot3 in a Gazebo scene and the corresponding LiDAR plot. We use these types of plots throughout this paper. We conduct experiments in a large room to test our algorithm on real-world data, and we place cardboard boxes to recreate the \ac{cfe}s generated by our method. An example of this real-world environment can be seen in \Cref{fig:case_2_real_world}.

The \ac{ml} model controlling the TurtleBot3 is a \ac{drl} model, trained using the Stable-Baselines3 implementation of the \ac{sac} algorithm \cite{stable-baselines3, haarnoja2018soft}. To help the agent adapt to new environments, we train the model using domain randomization \cite{tobin2017domain} to generate random environments in the Gazebo simulator \cite{koenig2004design_gazebo}. Our ROS package for generating these environments is available at \cite{sbremman2024turtlebot3}.

We formulate this as a \ac{rl} problem, extending our previous work in \cite{remman2024shapley}, where the goal is to navigate the TurtleBot3 to a specific location. We define the state space $\mathbf{s}$ as:

\begin{itemize}
    \item $\mathbf{s}_{\text{LiDAR}} = [s_1, s_2, \dots, s_{180}]$: A vector of 180 LiDAR distance readings collected at 2-degree intervals 360 degrees around the TurtleBot3,
    \item $\cos(\theta_g)$: The cosine of the angle to the goal,
    \item $\sin(\theta_g)$: The sine of the angle to the goal,
    \item $d_g$: The distance to the goal from the TurtleBot3.
\end{itemize}
All elements of $\mathbf{s}$ are normalized to lie within the range $[0, 1]$.

The action space $\mathbf{a} = [a_{\text{linear}}, a_{\text{angular}}]$ is defined as:
\begin{itemize}
    \item $a_{\text{linear}}$: The applied linear velocity of the TurtleBot3,
    \item $a_{\text{angular}}$: The applied angular velocity.
\end{itemize}
Both actions are scaled between $[-1, 1]$ using a hyperbolic tangent (tanh) activation function in the output layer, and multiplied by their respective maximum values for the TurtleBot3. These scaled commands are sent to the robot using messages to the \texttt{vel\_cmd} ROS topic.

The reward function we used, which we found using trial and error, is 

\begin{equation}
\begin{aligned}
R = & \, (-d_{\text{current}}) \\
& + \lambda_{\text{obstacle}} \cdot e^{-k \cdot \min(\mathbf{s}_{\text{LiDAR}})} \\
& + \lambda_{\text{backwards}} \cdot a_{\text{linear}},
\end{aligned}
\end{equation}

where:
\begin{itemize}
    \item $ d_{\text{current}} $ is the current distance to the goal,
    \item $ \lambda_{\text{obstacle}} = -2.0 $: Penalizes proximity to obstacles.
    \item $ k = 50.0 $: Controls the steepness of the exponential obstacle penalty.
    \item $ \lambda_{\text{backwards}} = 0.01 $: Penalizes backward movement.
    \item $ \mathbf{s}_{\text{LiDAR}} $ represents the vector of LiDAR readings, with $\min(\mathbf{s}_{\text{LiDAR}})$ as the distance to the closest obstacle,
    \item $ a_{\text{linear}} $ is the linear component of the action, with a penalty applied if moving \textit{backward}.
\end{itemize}
In addition, when the agent crashes or reaches the goal, it gets an additional reward of -100 or +100.

While traditional control methods could solve this navigation task, by using \ac{drl}, we create a good test case study for our \ac{cfe} generation algorithm. In addition, this task is a Partially Observable Markov Decision Process (POMDP), where solutions typically benefit from a memory of past states. However, for this case study, we simplify using only convolutional and feedforward layers, without recurrent elements, to simplify training and implementation and focus on testing the \ac{cfe} generation method.

To find optimal hyperparameters for training, we use Optuna \cite{optuna_2019} for hyperparameter tuning. The search space for \ac{sac} and the final values are shown in \Cref{tab:hyperparam_search_space} and \Cref{tab:optimized_hyperparams}, respectively.

Our neural network first processes the 180 LiDAR readings through two 1D convolutional layers using \textbf{circular padding}. The first layer has \textbf{4 output channels}, a \textbf{kernel size of 5}, \textbf{stride of 1}, and \textbf{padding of 2}. The second layer has \textbf{8 output channels}, a \textbf{kernel size of 5}, \textbf{stride of 2}, and \textbf{padding of 2}. The output is then combined with non-LiDAR inputs and passed through two fully connected layers with \textbf{128 units} each. We use ReLU activation functions, except for the tanh output layer.

\begin{table}
    \centering
    \caption{Hyperparameter search space for tuning the Stable-Baselines3 implementation of \ac{sac} using Optuna}
    \begin{tabular}{ll}
        \toprule
        \textbf{Hyperparameter} & \textbf{Range/Values} \\ 
        \midrule
        $\gamma$ & [0.9, 0.9999] (log) \\ 
        Learning Rate & [1e-4, 1e-2] (log) \\ 
        Buffer Size & [600,000, 1,000,000] \\ 
        Batch Size & [16, 400] \\ 
        $\tau$ & [1e-3, 1] (log) \\ 
        Entropy Coefficient & \{auto, 0.1, 0.01, 0.001, 0.05, 0.005\} \\ 
        Target Entropy & \{auto, -1, -2, -5, -10\} \\ 
        Use SDE & \{False, True\} \\ 
        Log Std Init & [-3, -0.5] \\ 
        \bottomrule
    \end{tabular}
    \label{tab:hyperparam_search_space}
\end{table}

\begin{table}
    \centering
    \caption{Optimized Hyperparameter Values}
    \begin{tabular}{ll}
        \toprule
        \textbf{Hyperparameter} & \textbf{Final Value} \\ 
        \midrule
        $\gamma$ & 0.9856 \\ 
        Learning Rate & 0.0009998 \\ 
        Buffer Size & 747,034 \\ 
        Batch Size & 220 \\ 
        $\tau$ & 0.1568 \\ 
        Entropy Coefficient & 0.01 \\ 
        Target Entropy & -2 \\ 
        Use SDE & True \\ 
        Log Std Init & -3.18 \\ 
        \bottomrule
    \end{tabular}
    \label{tab:optimized_hyperparams}
\end{table}

\section{Results and Discussion}\label{sec:results_discussion}

\subsection{Comparison with DiCE}\label{subsec:dice_comparison}

We compare our LiDAR \ac{cfe} generation methodology with the \ac{dice} method by Mothilal et al. \cite{mothilal2020dice}, using the Interpretml implementation \cite{dice2020github}. The version of \ac{dice} we use also uses a genetic algorithm for cost function optimization. We had to make alterations in the code to support multi-output regression models. For instance, we adapted the hinge loss described in \Cref{eq:hinge_loss} by summing the hinge losses across output nodes.

\begin{figure}
    \centering
    \begin{subfigure}[b]{1.0\columnwidth}
        \centering
        \includegraphics[width=\linewidth]{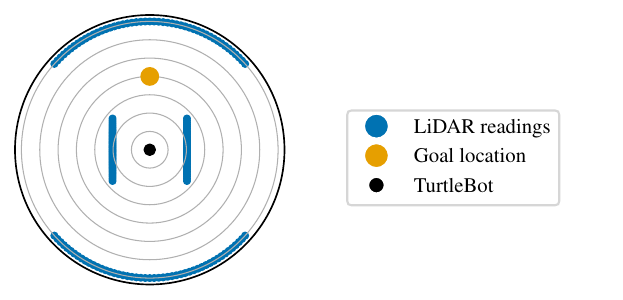} 
        \caption{Base LiDAR state}
        \label{fig:dice_comp_base_state}
    \end{subfigure}
    \begin{subfigure}[b]{0.5\columnwidth}
        \centering
        \includegraphics[width=\linewidth]{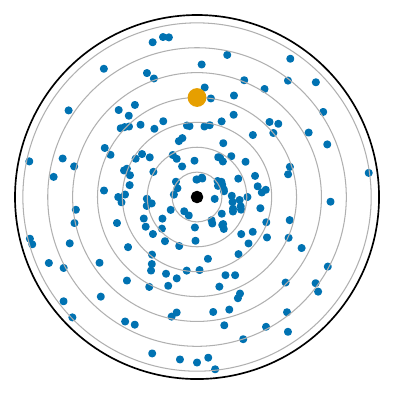}
        \caption{\ac{cfe} generated by \ac{dice}}
        \label{fig:dice_comp_dice}
    \end{subfigure}\hfill
    \begin{subfigure}[b]{0.5\columnwidth}
        \centering
        \includegraphics[width=\linewidth]{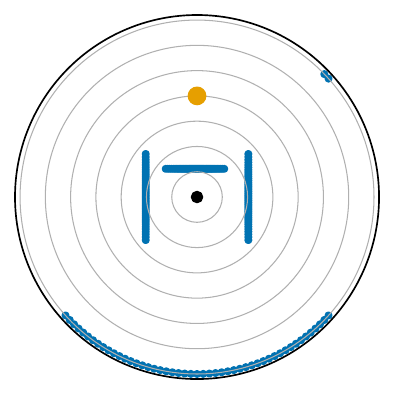}
        \caption{\ac{cfe} generated by our method}
        \label{fig:dice_comp_ours}
    \end{subfigure}
    \caption{Comparison between \ac{dice} and our algorithm on a simple case where the goal is right in front, with walls on either side.}
    \label{fig:dice_comp_whole_figure}
\end{figure}

We tested both methods in a scenario using generated LiDAR data where walls are positioned to the left and right of the TurtleBot3, with the goal ahead. The base state’s LiDAR readings are shown in \Cref{fig:dice_comp_base_state}. In this state, the policy applies the action\footnote{This action also causes a small left turn because the policy prefers to approach the goal from the left in this case.} $$\mathbf{a} = [a_{\text{linear}}, a_{\text{angular}}] = [0.9640, 0.4211].$$

To investigate what environmental changes would cause the agent to reverse, we ran both \ac{dice} and our algorithm with desired action bounds of $a_{\text{linear}} \in [-1.0, 0.0]$ and $a_{\text{angular}} \in [-0.1, 0.1]$. In this case, a rational \ac{cfe} could explain that a large obstacle between the agent and the goal would cause the agent to reverse. \Cref{fig:dice_comp_dice} shows a \ac{cfe} generated by \ac{dice}, and \Cref{fig:dice_comp_ours} presents one generated by our method. The \ac{dice} \ac{cfe} appears noisy, but our method produces a realistic environment. Adjusting proximity and sparsity weights in \ac{dice} to, in theory, increase the similarity to the base state still resulted in noisy \ac{cfe}s. Some possible reasons for \ac{dice}’s poorer performance include the high dimensionality of the LiDAR data (180 features), which our method works its way around by parameterizing the LiDAR space with the generated obstacles and exploiting spatial correlations between readings through the geometric objects. Although this comparison may arguably be biased towards our method, it highlights the value of domain knowledge in designing \ac{cfe} algorithms for specific data types. It could be interesting to try other optimization options within the \ac{dice} package to see if they produce more realistic \ac{cfe}s, but this is outside the scope of this paper and would require further adaptations for multi-output regression.

\subsection{Real-world experiments}

\subsubsection{Case 1: No Obstacles}

In this first test using real-world LiDAR data, we generate \ac{cfe}s for a scenario with no obstacles, and the goal is straight ahead. We define the counterfactual action as moving backward and run our algorithm with the parameters in \Cref{tab:no_obstacles_parameters}. Since similarity to the base state is not important here, we set $p_{\text{loss}}$ weight $\lambda_p = 0.0$ and use the \textit{min\_distance} combination function, which simplifies obstacle placement.

\begin{figure}
    \centering
    \includegraphics[width=.95\columnwidth]{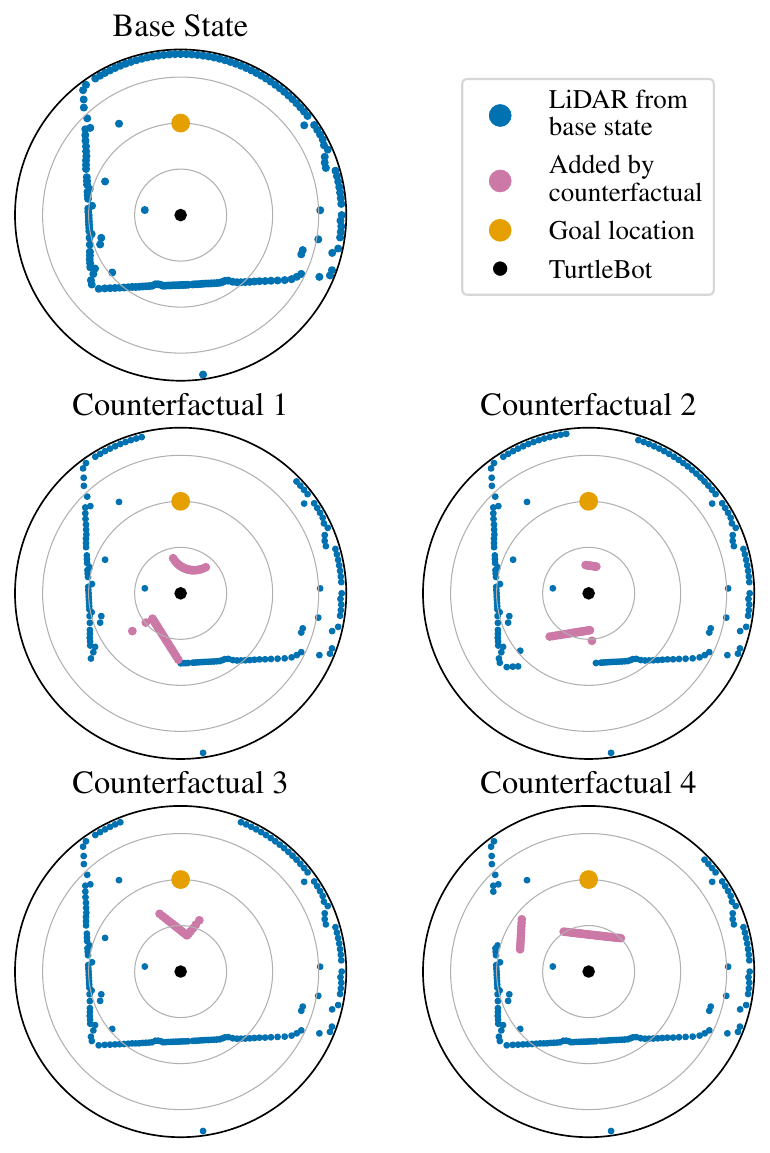}
    \caption{Four \ac{cfe}s generated by the algorithm from the base state in Case 1, with the desired action being to move backwards, and not turn much, which we define by $[a_{\text{linear}} \in [-1.0, 0.0]$ and $a_{\text{angular}} \in [-0.2, 0.2]$.}
    \label{fig:no_obstacles_cf_generated}
\end{figure}

We generate ten \ac{cfe}s and display four in \Cref{fig:no_obstacles_cf_generated}. Each shows that placing an obstacle directly in front of the TurtleBot3 leads the agent to choose to reverse, as expected. If obstacles only placed behind the agent could lead it to choose a reverse action, this could indicate irrational behavior and suggest a lack of model reliability.

\begin{figure}
    \centering
    \includegraphics[width=.9\columnwidth]{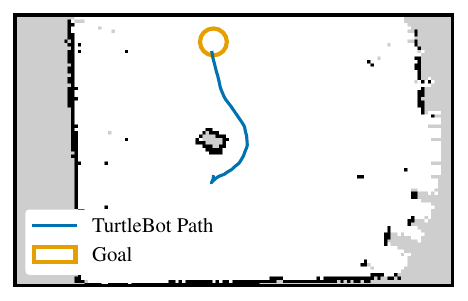}
    \caption{TurtleBot3 path on a map after adding an obstacle inspired by Counterfactual 3 in \Cref{fig:no_obstacles_cf_generated}.}
    \label{fig:no_obstacles_add_CF_physically_turtlebot_path}
\end{figure}

\begin{figure}
    \centering
    \includegraphics[width=.75\columnwidth]{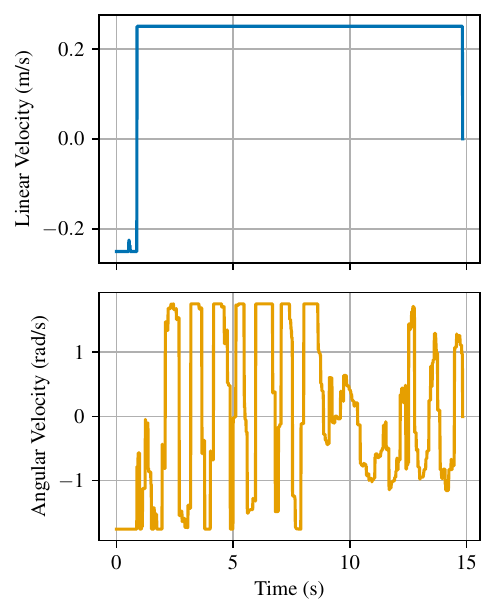}
    \caption{Actions chosen when we actualize counterfactual 3 in \Cref{fig:no_obstacles_cf_generated}.}\label{fig:no_obstacles_add_CF_physically_actions}
\end{figure}

To validate the generated \ac{cfe}s in \Cref{fig:no_obstacles_cf_generated}, we placed a cardboard box similar to the obstacle in Counterfactual 3. We then let the agent navigate to the goal and plot its path in \Cref{fig:no_obstacles_add_CF_physically_turtlebot_path} and chosen actions in \Cref{fig:no_obstacles_add_CF_physically_actions}. The TurtleBot3 begins by reversing, as the generated \ac{cfe} said. The angular velocity action seems noisy, likely due to differences between the simulation the agent was trained in and the real-world environment.

\begin{table}
    \centering
    \caption{Case 1, Algorithm Parameters and Results}
    \begin{tabular}{ll}
        \toprule
        \textbf{Parameter} & \textbf{Value} \\ 
        \midrule
        Time to Calculate & 205.51 s \\ 
        Combination method & min\_distance \\
        $a_{\text{linear}}$ desired bounds & [-1.0, 0.0] \\ 
        $a_{\text{angular}}$ desired bounds & [-0.2, 0.2] \\ 
        $y_{\text{loss}}$ weight, $\lambda_y$ & 1.0 \\
        $p_{\text{loss}}$ weight, $\lambda_p$ & 0.0 \\
        Number of obstacles generated & 5 \\
        \bottomrule
    \end{tabular}
    \label{tab:no_obstacles_parameters}
\end{table}

\subsubsection{Case 2: Adding Obstacles to Trick the Agent into Crashing}\label{subsubsec:case_2}

\begin{figure}
    \centering
    \includegraphics[width=.9\columnwidth]{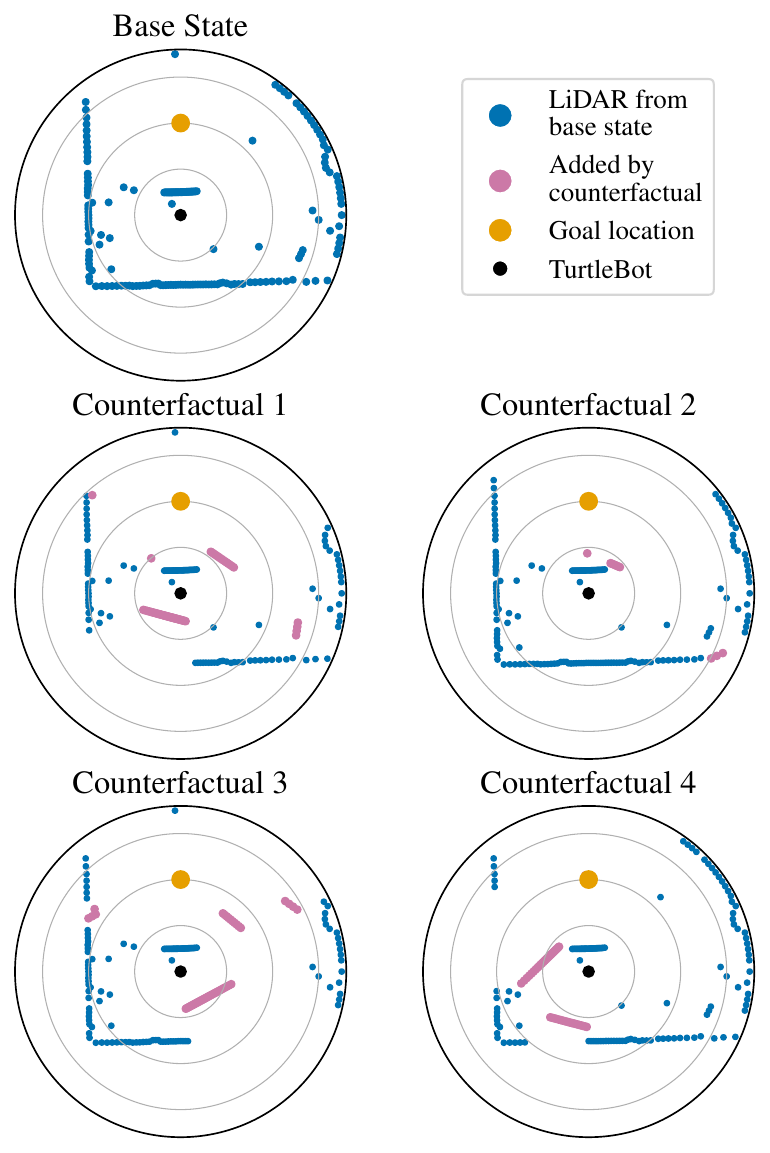}
    \caption{Four \ac{cfe}s generated by the algorithm from the base state in Case 2. We aim to make the policy drive the TurtleBot3 forward, which we define by $[a_{\text{linear}} \in [0.6, 1.0]$ and $a_{\text{angular}} \in [-1.0, 1.0]$.}
    \label{fig:obstacle_large_ahead_change_CF_counterfactuals}
\end{figure}

In the second case, we attempt to exploit the \ac{cfe}s to create a scenario where the policy crashes the TurtleBot3. We place a large cardboard box directly in front of the TurtleBot3 and use this as the base state. In this base state, the policy chooses to reverse to the left. To encourage the TurtleBot3 to move forward and potentially crash, we generate \ac{cfe}s for what it would take to make the agent drive forward instead. To generate the \ac{cfe}s, we use the parameters in \Cref{tab:obstacle_large_ahead_change_CF_parameters}. The base state and generated \ac{cfe}s are shown in \Cref{fig:obstacle_large_ahead_change_CF_counterfactuals}, and we select Counterfactual 2 for testing. \Cref{fig:case_2_real_world} shows pictures of how the base state and Counterfactual 2 looks in the real world.

\begin{figure}
    \centering
    \includegraphics[width=.9\columnwidth]{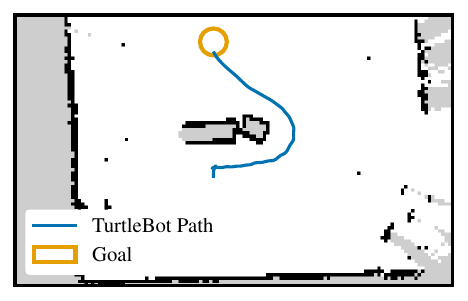}
    \caption{TurtleBot3 path on a map after adding an obstacle inspired by Counterfactual 2 in \Cref{fig:no_obstacles_cf_generated}. Notice the two obstacles in the middle of the map, which are the two cardboard boxes added.}
    \label{fig:obstacle_large_ahead_turtlebot_path}
\end{figure}

Counterfactual 2 suggests adding more obstacles in front of the TurtleBot3, which seems not to be logical for causing the agent to drive forward. We place an additional small box slightly to the right of the larger box and observe the resulting path in \Cref{fig:obstacle_large_ahead_turtlebot_path}. Although the TurtleBot3 initially moves forward, it avoids the obstacles, making slight maneuvers forward and backward before turning right and navigating around them. This illustrates a limitation of our method and \ac{cfe}s in general for explaining sequential decision processes: an agent’s action in one step may not predict its subsequent behavior.

\begin{table}
    \centering
    \caption{Case 2, Algorithm Parameters and Results}
    \begin{tabular}{ll}
        \toprule
        \textbf{Parameter} & \textbf{Value} \\ 
        \midrule
        Time to Calculate & 39.84 s \\ 
        Combination method & min\_distance \\ 
        $a_{\text{linear}}$ desired bounds & [0.6, 1.0] \\ 
        $a_{\text{angular}}$ desired bounds & [-1.0, 1.0] \\ 
        $y_{\text{loss}}$ weight, $\lambda_y$ & 1.0 \\
        $p_{\text{loss}}$ weight, $\lambda_p$ & 0.0 \\
        Number of obstacles generated & 5 \\
        \bottomrule
    \end{tabular}
    \label{tab:obstacle_large_ahead_change_CF_parameters}
\end{table}

\begin{figure}
    \centering
    \includegraphics[width=.75\columnwidth]{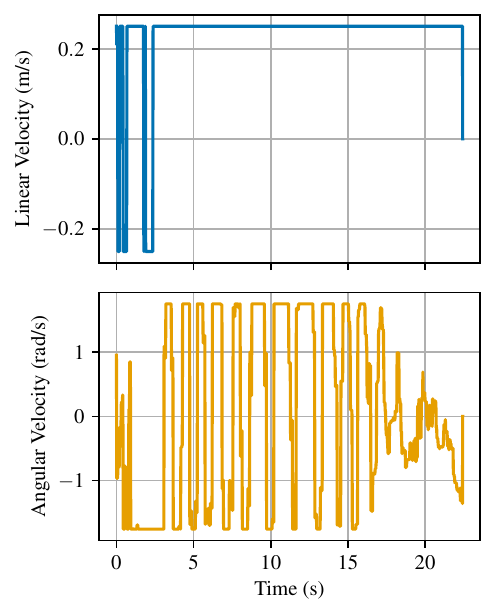}
    \caption{Actions chosen when we actualize counterfactual 2 in \Cref{fig:no_obstacles_cf_generated}.}\label{fig:obstacle_large_ahead_actions}
\end{figure}

\subsection{Case 3: Explaining the Policy's Left-Turn Preference}

\begin{table}
    \centering
    \caption{Case 3, Algorithm Parameters}
    \begin{tabular}{ll}
        \toprule
        \textbf{Parameter} & \textbf{Value} \\ 
        \midrule
        Combination method & min\_distance \\ 
        $a_{\text{linear}}$ desired bounds & [0.9, 1.0] \\ 
        $a_{\text{angular}}$ desired bounds & [-1.0, -0.5] \\ 
        $y_{\text{loss}}$ weight, $\lambda_y$ & 1.0 \\
        $p_{\text{loss}}$ weight, $\lambda_p$ & 0.1 \\
        Number of obstacles generated & 1 \\
        \bottomrule
    \end{tabular}
    \label{tab:model_understanding_parameters}
\end{table}

\begin{figure}
    \centering
    \includegraphics[width=.9\columnwidth]{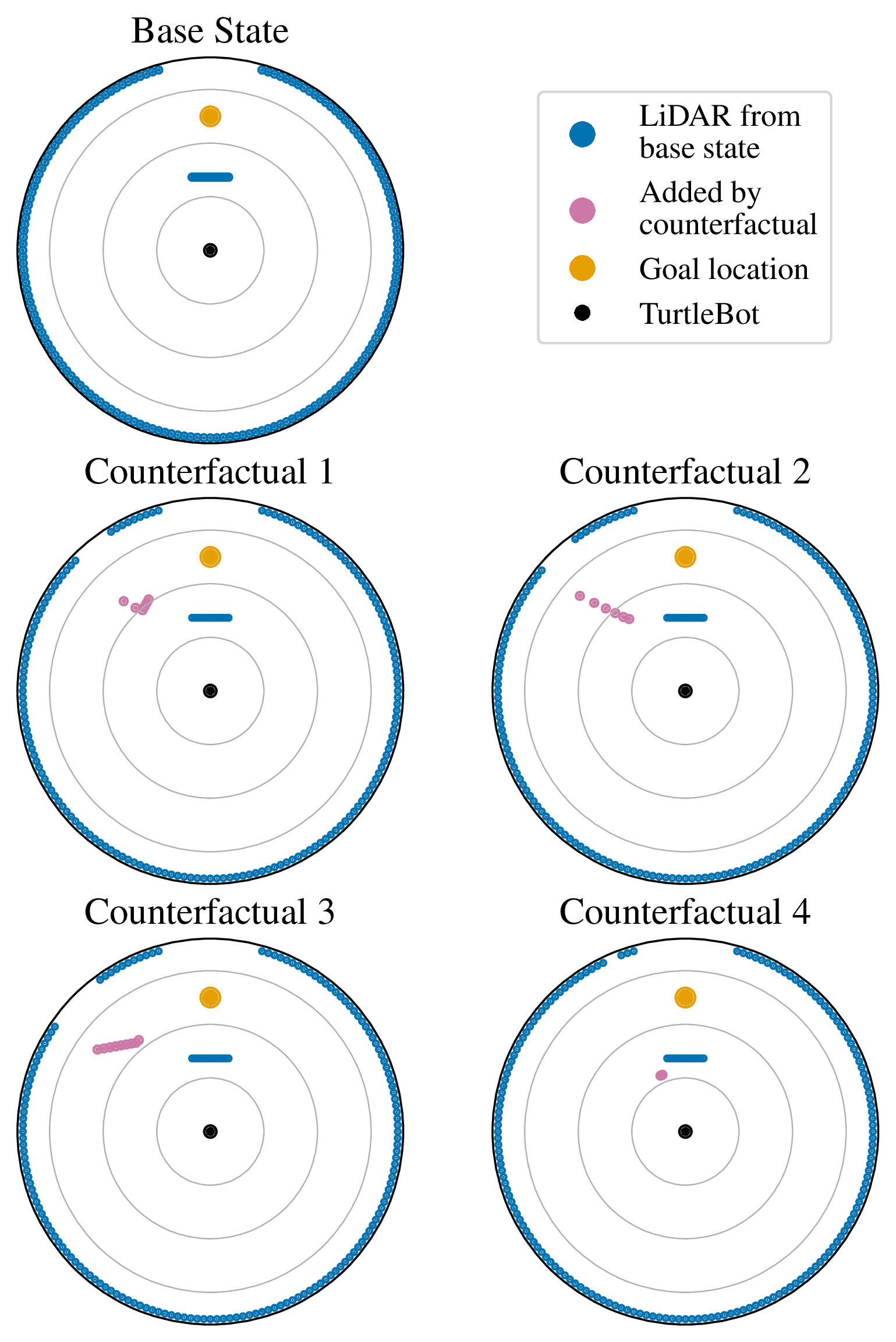}
    \caption{Base state and \ac{cfe}s for Case 3}\label{fig:model_understanding}
\end{figure}

After observing that our \ac{drl} policy often prefers to turn to the left, we try to gain an understanding of this by generating \ac{cfe}s. Using virtual data, we simulate a base LiDAR state with a box obstacle 2.5 meters in front of the LiDAR origin (shown in the top left of \Cref{fig:model_understanding}), where the policy selects the action 
$$\mathbf{a} = [a_{\text{linear}}, a_{\text{angular}}] = [0.9640, 0.9640].$$

We run our algorithm with the parameters in \Cref{tab:model_understanding_parameters} and generate 100 \ac{cfe}s, selecting $\lambda_p = 0.1$, to ensure more similarity to the base state. Four of the generated \ac{cfe}s are shown in \Cref{fig:model_understanding}. In all 100 generated \ac{cfe}s, there where only obstacles on the left, even when we set $\lambda_p = 0$ as in the other cases. This result shows that the "base decision" for the agent in this case is to turn to the left and would only not do so if there was an obstacle in the way. Even though these explanations might seem obvious, obvious explanations demonstrate our methodology's strengths and that the agent behaves, at least to some extent, according to human intuition. In this case, we could interpret the user's question and method's explanation in natural language as: 

\begin{quote}
User: "Why don't you turn to the \textit{right} here?"\\
Agent: "I would turn to the \textit{right} instead of to the \textit{left} if there was an obstacle on the \textit{left}."
\end{quote}
\vspace{-0.1cm}
\section{Conclusion}\label{sec:conclusion}
In this paper, we introduced a method for generating realistic \glspl{cfe} for \ac{ml}-controlled mobile robots using 2D LiDAR data to make them more interpretable. Our approach uses genetic algorithms and domain-specific insights to produce realistically structured LiDAR \glspl{cfe} that are intuitive and give us a good understanding of the \ac{ml} model's control decisions.

We perform experiments with a TurtleBot3 controlled using \ac{drl} in both simulated and real-world environments to demonstrate how the method can generate realistic \glspl{cfe} that improve understanding of autonomous control behavior. This work contributes to advancing explainable AI in mobile robotics, and our method can serve as a tool for enhancing transparency and reliability in both \ac{ml}-based and traditional control systems. Future work could extend this approach to 3D LiDAR, other controllers, and other robotic applications. 

\section*{Acknowledgment}
The Research Council of Norway supported this work through the EXAIGON project, project number 304843.

\bibliographystyle{IEEEtran}
\bibliography{mylib}

\end{document}